\documentclass[journal]{IEEEtran}
\usepackage{amsmath,mathtools,amsfonts,amssymb,amsthm}
\usepackage{xcolor}
\usepackage{cite}

\usepackage{color}

\usepackage[draft]{minted}
\usepackage{tcolorbox}
\usepackage{algorithm}
\usepackage{algpseudocode}

\usepackage[compact]{titlesec}         
\titlespacing{\section}{0pt}{0pt}{0pt} 
\AtBeginDocument{
  \setlength\abovedisplayskip{0pt}
  \setlength\belowdisplayskip{0pt}}

\ifCLASSINFOpdf
\else
\fi

\begin{document}
%
\title{OpenGridGym: An Open-Source AI-Friendly Toolkit for Distribution Market Simulation}
%
%
%

\author{Rayan~El~Helou$^\ast$,~Kiyeob~Lee$^\ast$,~Dongqi~Wu$^\ast$,\\
Le~Xie$^{\ddagger,~\S}$,~Srinivas~Shakkottai$^\dagger$~and~Vijay~Subramanian$^\dagger$\\
        \{$^\ast$\IEEEmembership{Student Member},
        $^\dagger$\IEEEmembership{Senior Member},
        $^\ddagger$\IEEEmembership{Fellow}\},~IEEE
\thanks{Rayan El Helou, Kiyeob Lee, Dongqi Wu, Le Xie and Srinivas Shakkottai are all with the Department of Electrical and Computer Engineering, Texas A\&M University, College Station, TX, USA.}
\thanks{Vijay Subramanian is with the Department of Electrical and Computer Engineering, University of Michigan, Ann Arbor, MI, USA.}
\thanks{$\S$ Corresponding Author: Le Xie, email: le.xie@tamu.edu}
}

%
%

\maketitle

\begin{abstract}
This paper presents OpenGridGym,  an open-source Python-based package that allows for seamless integration of distribution market simulation with state-of-the-art artificial intelligence (AI) decision-making algorithms. 
We present the architecture and design choice for the proposed framework, elaborate on how users interact with OpenGridGym, and highlight its value by providing multiple cases to demonstrate its use. {\color{black} Four modules are used in any simulation: (1) the physical grid, (2) market mechanisms, (3) a set of trainable agents which interact with the former two modules, and (4) environment module that connects and coordinates the above three. We provide templates for each of those four, but they are easily interchangeable with custom alternatives.} Several case studies are presented to illustrate the capability and potential of this toolkit in helping researchers address key design and operational questions in distribution electricity markets. 
\end{abstract}


%
\IEEEpeerreviewmaketitle

\section{Introduction}
%
%
%
\IEEEPARstart{M}{odern} electric grids are shifting from a centralized  to a more distributed architecture. This brings up a new set of operational challenges due to the expected rapid growth of a diverse set of distributed energy resources (DERs), such as rooftop photovoltaics (PVs), electric vehicles (EVs), and storage systems at the grid edge. It also introduces more decision-makers (agents) into the picture who could strategically game the system under decentralized electricity market designs. Adding decision-makers such as DER owners, flexible loads and aggregators may significantly influence both physical and market operations. Thus, it is indispensable to understand in modern grids the implications of different market design and operational issues.

In particular, distribution grids are more susceptible to such evolution than transmission grids, as the latter is more capable at dampening intermittencies in load and generation than the former. It is necessary to focus more of our efforts on the design of distribution-level markets, and to explore alternatives to traditional approaches that rely on outdated assumptions of the distribution grid. There is a wide range of candidate market mechanisms for modern distribution grids, primarily due to the fact that much more decision-makers are involved, each with more constraining requirements than those of a transmission-level aggregator.

For example, conventional electric consumers at the distribution grids are fixed rate payers with the rate determined a-priori by the regulatory agencies. Local electric utilities plan the distribution grid capacities accordingly based on expected load growth. However, such a mechanism may be rendered ineffective due to the lack of clear incentives to encourage time-varying load flexibility at the operational stage.  To address the need for new institutional design at the distribution grid level, alternatives to this pricing mechanism have been proposed in \cite{athindra} where prices dispatched to \textit{prosumers} (i.e. two-way power usage) are driven by their consumption patterns over time, measured using smart meter data. Nonetheless, demand in the short run is assumed to be inelastic. Similar market mechanisms for a wholesale market-like distribution locational marginal pricings (DLMPs) have also been introduced and conceptually investigated in the literature \cite{caramanis_transformer}.

While there is a growing body of literature that advocates the use of DLMPs and theoretical properties of DLMPs have been investigated \cite{dlmp_voltage, dlmp_fairness, bose_dlmp}, concrete comparisons with alternatives to DLMPs are missing and are yet to be analyzed both theoretically  and empirically. This is an important gap between the conceptualization of how market clearing and pricing should be done via DLMPs and its implementation in practice. Moreover, while game theory offers analytical tools to investigate strategic interactions on how DER, DER aggregators and flexible demands may participate in future electricity markets, without making substantial assumptions on how decision-makers strategically interact, analysis of game theoretic models is intractable in many settings  \cite{bose2019some}.

{\color{black}Similar to the LMP calculation in bulk transmission system market operation using Optimal Power Flow (OPF), the approach and formulation have been proposed for the computation of DLMP in \cite{low_opf} \cite{low_exact} for distribution networks, using the DistFlow power flow model for radial networks, with the objectives to minimize either power loss or generation cost.} In reference \cite{bose_dlmp}, the properties of convex relaxations using the DistFlow model are investigated, and Distribution Locational Marginal Pricing (DLMP) is formulated, and results reveal the nature of DLMP distribution relative to physical constraints. In \cite{caramanis_transformer} the degradation cost of transformers is explicitly factored in the LMP calculation for distribution networks.

{\color{black} The inelastic nature of electricity demand is frequently the binding constraint of LMP-based mechanisms in OPF-based power system planning, where many ultra-high marginal price result from the necessity to meet invariant local demand either by using expensive reserve or sub-optimal dispatch. Demand response (DR) programs can incentivize end-users in real-time to adapt their power consumption to the availability of electric power generation and delivery and introduce elasticity.} References \cite{ming2020prediction}, \cite{xia2017energycoupon} and \cite{li2015energy} propose a novel DR program where end-users receive coupons as incentives to shift power consumption from peak hours to off-peak hours. This approach has been tested by conducting a case study on real users in Texas, with results that suggest that people respond positively to such incentives. Such programs can be easily implemented using our proposed simulation framework.

\begin{figure}[!ht]
\vspace{-0.5em}
\centering
\includegraphics[width=0.3\textwidth]{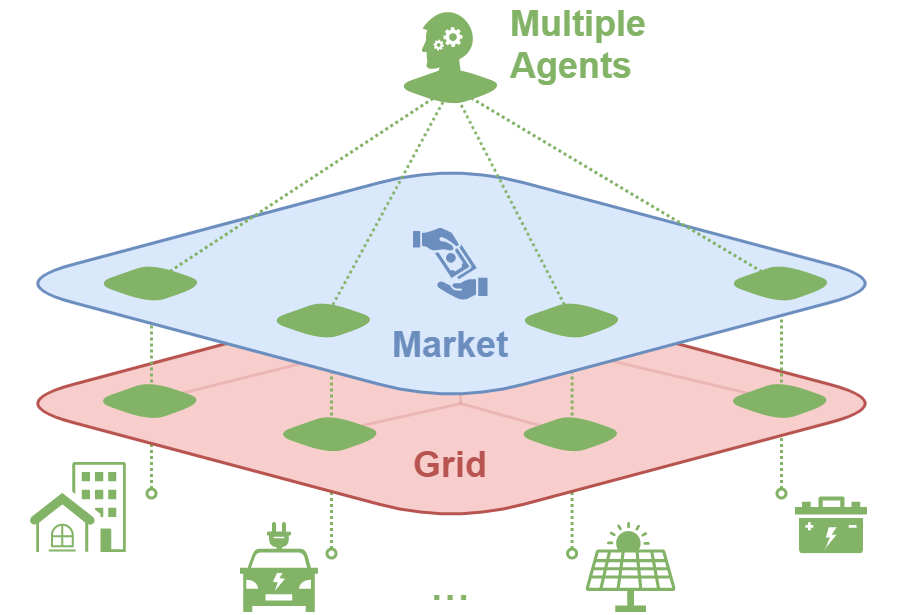}
\caption{Envisioned distribution grids with DERs participating in local markets.}
\label{fig:layers}
\vspace{-0.5em}
\end{figure}



\subsection{Existing Approaches to Electricity Market Simulation}
There already exists a set of widely-used open-source tools for simulating physical operations of both transmission grids (e.g. MATPOWER \cite{MATPOWER} and pandapower \cite{pandapower}) and distribution grids (e.g. OpenDSS \cite{OpenDSS} and GridLAB-D \cite{GridLABD}), and a proposed set of tools for electricity market simulation (e.g. AMES test bed \cite{AMESTestBed}). All these tools work well under a traditional assumption of \textit{weak coupling} between physical and market models of electric power grids. This assumption needs to be re-visited in modern grids due to the intermittence of electricity generation and consumption, as well as the likely participation of a variety of large and small agents.

In this paper, we provide a simulation toolkit that can help researchers simulate and compare the outcomes of various market mechanisms for realistic distribution grids. As implied in Figure \ref{fig:layers}, we rely on a framework that enables modular representation of grids, markets, and DER-controlling agents, which could potentially participate either in market or physical grid operations. In contrast, the existing set of tools is either not friendly to learning-based algorithms, or does not provide an easy-to-interchange modular structure which enables experimenting with various models for both grid and market operations.


\subsection{Our Contributions}
To account for the multiplicity of agents in distribution electricity markets, we propose a new formulation where demand and supply entities are agents that submit bids or offers into the market, and prices are dispatched to them by a market operator. We develop an OpenAI Gym-like \cite{Gym} benchmark for testing such multi-agent environments and invite machine learning (ML) and power systems experts to test the use of reinforcement learning (RL) to fulfill user-defined objectives. A similar effort has been made by RTE France with their Grid2Op platform, an environment popularized by the L2RPN competition \cite{L2RPN}, but the platform is restricted to physical operations, and does not include market operations. Reference \cite{Glavic2019} reviews the use of RL in electric power systems, and the aim in our proposed work is to provide a benchmark environment for distribution electricity market simulations.

Here are the key contributions of our work:
\begin{itemize}
    \item We propose a framework to serve as a benchmark for joint market and distribution grid simulation in a competitive multi-agent setting.
    \item We provide an open-source Python-based user-friendly toolkit for performing simulations with trainable AI-driven agents, with use cases to demonstrate it.
\end{itemize}

The remainder of this paper is organized as follows. In Section \ref{sec:OpenGridGym}, we introduce our proposed Python-based package, OpenGridGym \cite{opengridgym_github}, and we provide a high-level overview of its architecture and user interface. In Section \ref{sec:Modules}, we dive deeper into the definitions of each simulation module and how those modules interact with one other to deliver a user-friendly experience. In Section \ref{sec:Use_Cases}, we present three different use cases of OpenGridGym that showcase its extensibility. In Section \ref{sec:Conclusion}, we provide concluding remarks and future directions.

\vspace{1em}
\section{OpenGridGym}
\label{sec:OpenGridGym}


\subsection{Motivation}
As distribution grids continue to modernize, the need to reconsider traditional market mechanisms grows, and some questions need to be raised. For example, should local markets be based on peer-to-peer transactions, or should the distribution grid be governed by DLMP-based mechanisms, and what are the implications of such propositions? To answer these questions in a consistent manner, we should be able to easily integrate models of physical distribution grids with market models in a unified framework which is accessible to all researchers in this domain. Moreover, what role does artificial intelligence (AI) play in shaping the future of electricity markets? By using a programming language like Python, an entire ecosystem of AI-friendly tools is inherently available to a researcher who seeks to answer such questions.

With our proposed modular Python-based simulation package, OpenGridGym \cite{opengridgym_github}, it is easy to swap out market mechanisms while keeping the same physical grid model, and vice-versa, to answer questions such as the ones asked above. The name ``OpenGridGym" refers to the fact that we provide an open-source environment to exercise various grid and market designs, and to train agents to meet their objectives. By leveraging existing open-source simulation tools already developed by researchers (e.g. OpenDSS by EPRI \cite{OpenDSS}), OpenGridGym is able to account for new modalities such as real-time monitoring of substations, distributed storage, photovoltaics, electric vehicles, and more. Off-the-shelf AI-friendly tools for learning and optimization, such as PyTorch \cite{PyTorch} and CVXPY \cite{CVXPY}, are readily available to the user, unlike with existing non-Python-based market simulators. We provide use cases in Section \ref{sec:Use_Cases} to illustrate the use of OpenGridGym.


\subsection{Proposed Architecture and Design Decisions}
In this subsection, we present the high-level architecture design of our simulation framework, and in the section that follows, we go deeper into how each of its components (\textit{Grid}, \textit{Market} and \textit{Agents}) operate individually and collectively.

\begin{figure}[!ht]
\centering
\includegraphics[width=0.49\textwidth]{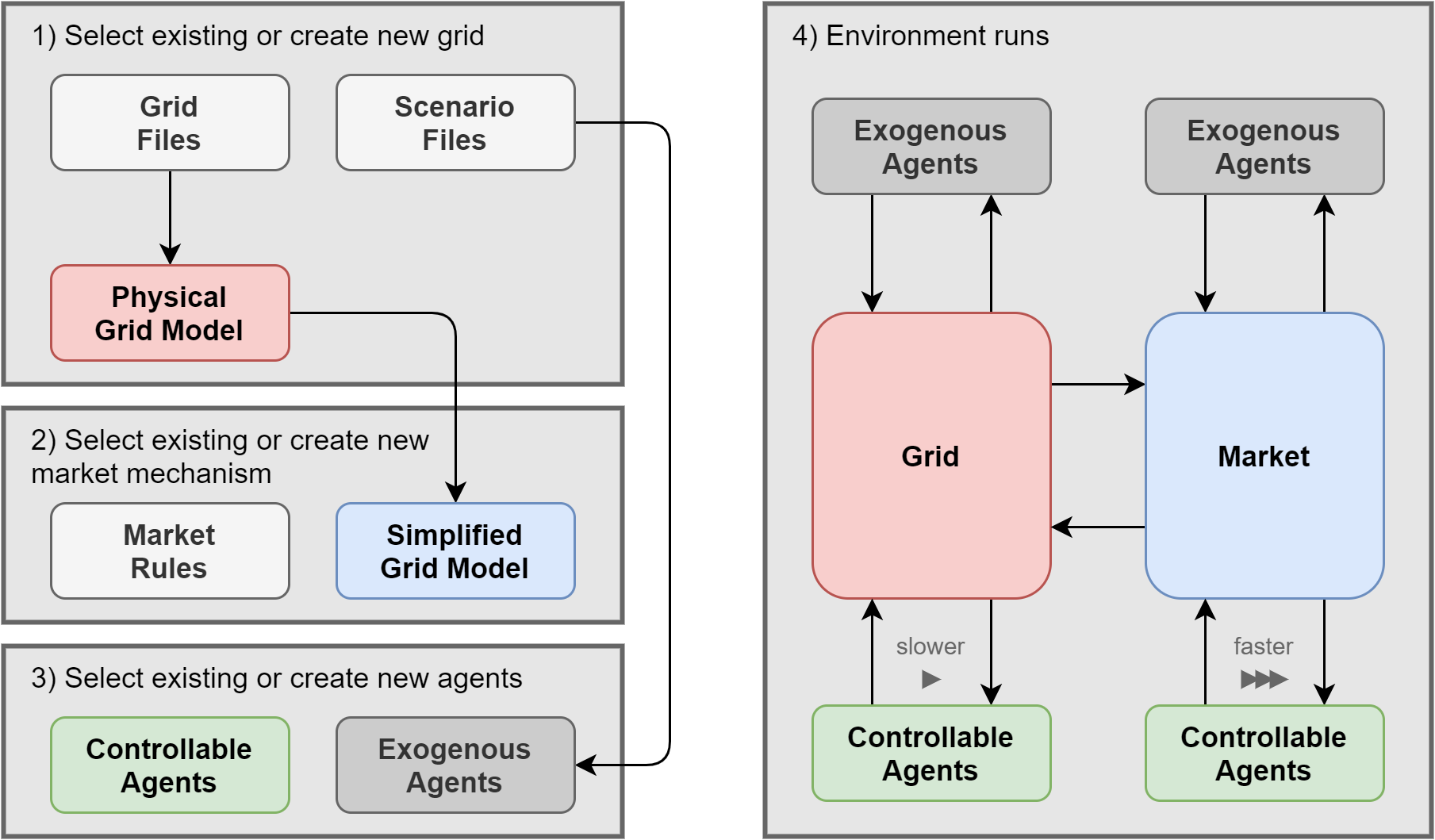}
\caption{User interface and simulation flowchart. The terms \textit{slower} and \textit{faster} indicate that market negotiations occur much more frequently than post-market-clearing interactions with the grid.}
\label{fig:architecture}
\end{figure}

The goal of this tool is to introduce a framework for simulating electricity markets. We suggest the use of Python as the main programming language partly because it is open-source, but mainly because it hosts an ecosystem of AI- and ML-friendly tools which are readily available to the user. Our work is inspired by the well-known OpenAI Gym \cite{Gym}, which facilitates the simulation of Markov Decision Processes (MDPs) and testing reinforcement learning (RL) algorithms. OpenAI Gym provides the user with two main contributions: 1) A set of (empty) \textit{base classes} which the user can fill in to represent their simulation environment, and 2) a set of use cases to demonstrate multiple examples of how such \textit{base classes} can be filled in by the user to provide meaningful outcomes. We aim to contribute similarly in our work.

In the section which follows, we introduce the Python-based \textit{base classes} for simulating electricity grids, market mechanisms, and various agents' behaviors which influence the two. Our design philosophy is the following. Each user of OpenGridGym should be given as much flexibility as possible over what they'd like to simulate, while simultaneously maintaining a level of consistency in the framework under which all researchers would operate. The implications are twofold. First, there needs to be minimal constraints on users' choice of grid format, market mechanism, and agents' behaviors. Second, we provide entirely Python-based easy-to-edit \textit{base classes} which act as building blocks for the user to implement their own version of a market simulation. Additionally, we do provide a set of templates, showcased in the Section \ref{sec:Use_Cases}, where we demonstrate how OpenGridGym could be used.

\subsection{User Interface}
As labelled on the left of Figure \ref{fig:architecture}, there are three steps the user needs to follow to begin the simulation process (to create a grid, a market, and agents). Each of these steps involve creating their own or selecting from existing modules. The order specified in the figure is based on the principle that the physical grid is lower-level than the market which acts on top of it. Here is the order:
\begin{enumerate}
    \item The user selects a grid from some case file or folder which contains the full detailed model of the physical system (indicated by the \textit{Physical Grid Model} block). This implicitly means that the user also selects which simulator they prefer to use (e.g. OpenDSS \cite{OpenDSS}). The \textit{Scenario Files} block indicates any exogenous input that might affect the grid state, e.g. weather and load data. Hence, it points to the \textit{Exogenous Agents} block.
    \item The user selects or creates a module to represent the market mechanism. For each market mechanism in general, there is a set of rules which dictate the prices dispatched to all participants. More specifically for electricity markets, since distribution (or transmission) grids are physical systems with constraints that need to be obeyed, any responsible market mechanism should also somehow model the grid to take those constraints into account. Such a model of the physical grid could be simple for market clearing purposes. This is indicated by the \textit{Simplified Grid Model} block in the figure.
    \item The user initializes a list of agents in the environment. Those agents are each an object in Python and they do not need to share the same decision-making policies. For each agent listed, the user either selects or creates a module to capture how they interact with either the physical grid directly or just with the market. Here is where Python's AI-friendly ecosystem can be utilized. For example, the \textit{Controllable Agents} block refers to the fact that agents are not parametrized by a predetermined set of files (e.g. the \textit{Scenario Files}). Rather, a learning algorithm can be associated with them so that each agent can individually seek its personal objective. This implies that such agents are effectively players in a stochastic game provided that there's more than one of them.
\end{enumerate}

\begin{figure}[!ht]
\centering
\includegraphics[width=0.35\textwidth]{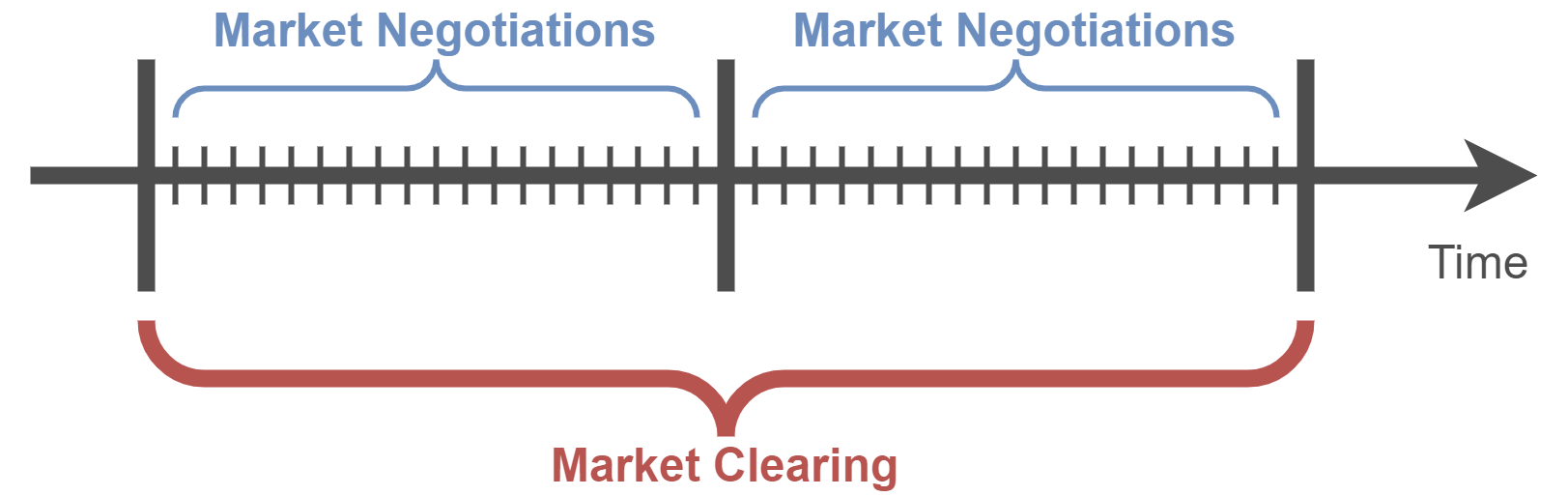}
\caption{Two-timescales for agent participation, one for negotiating in the market, and one for acting on the physical grid after the market clears.}
\label{fig:two_timescales}
\end{figure}

Once those three steps are complete, the environment can run as shown in the fourth step of the figure. We employ a two-timescale discrete-time sequential process which simulates the interplay between the metaphysical market and the physical grid. This is illustrated in Figure \ref{fig:two_timescales}. The exact formulation of this process is shown in Algorithm \ref{alg:environment} in the section that follows, but simply put, negotiations in the marketplace occur much more frequently than changes to the physical grid which are governed by market clearing. Our philosophy behind this is the following. The physical grid is governed at every time step by choices made by individuals who are faced with various opportunities, namely to increase or decrease electricity consumption (or production). We assume that those choices are determined as a result of negotiations made in a marketplace. For this simple reason, we model the market with a faster time-scale. To be clear, when we say one step is taken in the physical grid, we refer to time scales closer to those of tertiary control in power grids, not to those of primary and secondary control. Transient control can still occur in between time steps marked by the term \textit{Market Clearing} in Figure \ref{fig:two_timescales}, but we do not impose any restrictions on such control in our proposed framework.

\vspace{1em}
\section{Python-based Modules}
\label{sec:Modules}
In this section, we define all the modules involved in the simulation process and explain how they tie together to provide a reasonable user experience. There are four modules involved in any instantiation of our proposed framework, and for each of those, we will describe the concept and show some minimal snippets of Python code to illustrate the ease of their implementation. The four modules are \textit{Grid}, \textit{Market}, \textit{Agent} and \textit{Environment}, as discussed earlier at a higher level in the previous section.

\begin{figure}[!ht]
\centering
\includegraphics[width=0.35\textwidth]{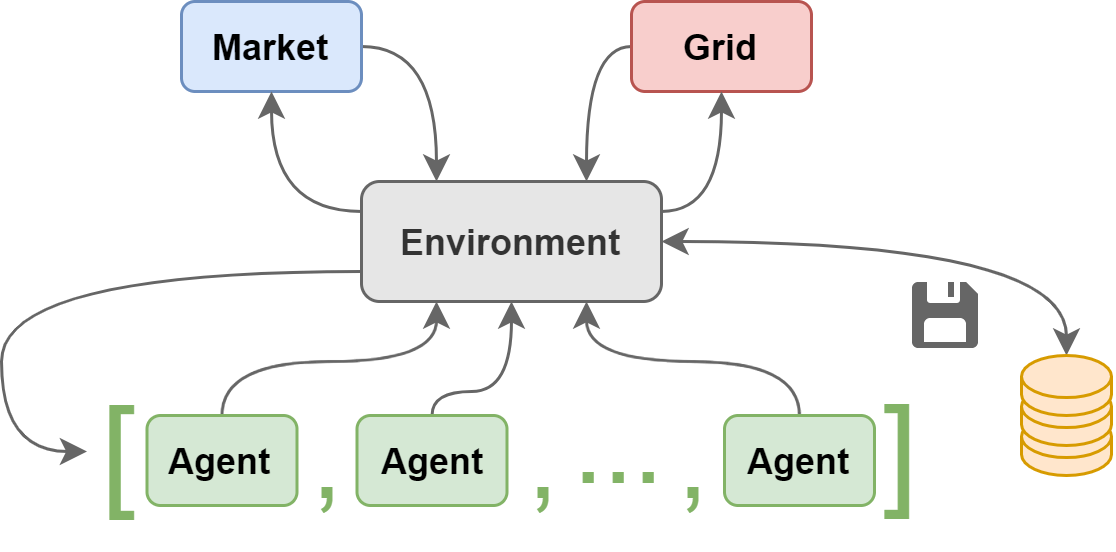}
\caption{Implicit access of Python objects to one another via the \textit{Environment} object. The head of an arrow is accessible as an attribute by the object on the tail of the arrow. Note: agent-to-agent communication is inherently possible.}
\label{fig:communication}
\end{figure}

In contrast with OpenAI Gym's framework \cite{Gym} for modelling discrete-time sequential processes, we do not propose that agents submit actions directly to the \textit{Grid} object and receive observations or rewards in return. Rather, each agent sets their actions in their local memory, and the environment is expected to pull them as needed. OpenAI Gym's approach works well in single-agent settings with one environment module involved. However, when there are several players (\textit{Grid}, \textit{Market} and multiple \textit{Agents}), building an explicit communication layer is cumbersome.
Our simple solution is that each of those players or objects has direct access to an object we call the \textit{Environment}, which in turn can access all other objects, as shown in Figure \ref{fig:communication}. One of the implications of this is that all objects can implicitly talk to one another, thanks to Python's built-in pointer system.

Off-the-shelf reinforcement-learning (RL) packages can still be used to train agents under this new framework. For each agent, the user would just need to ensure that they can extract actions, observations and rewards at any time step. For example, for an agent to access observations, say voltages, from the grid, we can simply write \texttt{agent.env.grid.get\_voltages()} in Python. Similarly, for the \textit{Market} object to receive actions from some agent, the user simply calls \texttt{market.env.agent.get\_market\_actions()}.

There is no strictly correct way to model any of the four objects introduced in the following subsections, as argued in the previous section. However, we propose that if the user follows the template we provide, then they can more consistently control experiments to compare market mechanisms with one another, agent behaviors with one another, and grid models with one another.

\subsection{Grid}
We propose that each grid object the user instantiates must have at least two functions implemented: \texttt{reset} and \texttt{step}. This is shown in the snippet code below.

\vspace{0.75em}
\begin{tcolorbox}
\begin{minted}[fontsize=\footnotesize]{python}
class CustomGrid(BaseGrid):
    def __init__(self, dss_case='',
                      scenarios=''):
        self.dss = DSS(case=dss_case)
        self.scenarios = scenarios
        ...
    def reset(self):
        self.t = -1
        ...
    def step(self):
        ...
        self.dss.solve()
        ...
\end{minted}
\end{tcolorbox}
\vspace{0.75em}

At the beginning of each episode, the environment automatically calls \texttt{env.grid.reset()} to reset the state of the grid. This includes resetting timing, weather information, any control equipment's states, scenarios, etc. At every iteration, the environment automatically calls \texttt{env.grid.step()} to execute all agents' actions, to solve power flow, to update its states and to store any relevant information as a result of agent's actions.

The actual model of the grid, labelled as \texttt{CustomGrid} in the snippet of code above, is either created (and filled in) by the user or selected from existing templates which we provide. For example, for a use case we show later in this paper, we rely on an IEEE 34-bus distribution grid in which the \texttt{step} function collects agents' actions and uses those to update the OpenDSS-based model accordingly.

Under this framework, leaving the \texttt{step} function empty is equivalent to simulating scenarios where agents do not influence the physical grid at all. This could be useful when experimenting with market-only interactions.

\subsection{Market}
Similar to the grid object, the market object also expects two functions to be implemented by the user: \texttt{reset} and \texttt{step}.

\vspace{0.75em}
\begin{tcolorbox}
\begin{minted}[fontsize=\footnotesize]{python}
class CustomMarket(BaseMarket):
    def __init__(self):
        ...
    def reset(self):
        self.t = -1
        ...
    def step(self):
        ...
\end{minted}
\end{tcolorbox}
\vspace{0.75em}

That is, there is no restriction on how the market is modelled, provided that the user specifies in the \texttt{reset} function how the market initializes any states or information it may derive based on the grid, and that the user specifies in the \texttt{step} function how the market uses agents' actions to dispatch electricity prices and quantities to be consumed or produced by all participants.

As shown in Algorithm \ref{alg:environment}, this market object iterates through a sequential negotiations process, but at no point in the process is the physical grid affected. Once the market negotiations terminate, we declare that the market has cleared. Based on this, agents can then determine the actual amount to consume or produce as described in the following subsection.

\subsection{Agent}
The minimal requirements on agents is that they are able to state what actions they'd like to apply on the market and on the grid. During the environment's market updates, agents' actions are expected to be declared in the \texttt{set\_market\_actions} function. Similarly, during grid updates, agents' actions are expected to be declared in \texttt{set\_grid\_actions}.

\vspace{0.75em}
\begin{tcolorbox}
\begin{minted}[fontsize=\footnotesize]{python}
class CustomAgent(BaseAgent):
    def __init__(self):
        ...
    def set_market_actions(self):
        ...
    def set_grid_actions(self):
        ...
\end{minted}
\end{tcolorbox}
\vspace{0.75em}

That is, by implementing those two functions for some agent, users would have completely modelled the behavior of said agent. Of course, those functions could internally rely on other local functions, which the user is free to create in Python. We provide examples of agents in the use cases. One example is a producer agent which submits actions in the form of supply curves that the market uses to determine optimal price and quantity dispatch.

\subsection{Environment}
Finally, the environment object, which ties the \textit{Grid}, \textit{Market} and \textit{Agent} objects together is presented here. We illustrate this in Algorithm \ref{alg:environment}.  
\begin{algorithm}
\caption{Environment Episode}\label{alg:environment}
\begin{algorithmic}[1]
\State Instantiate the following objects in Python:
\begin{itemize}
    \item \texttt{grid} \Comment{e.g. with connection to OpenDSS}
    \item \texttt{market} \Comment{constrained by simplified grid model}
    \item \texttt{agents} \Comment{with scenarios and user profiles}
\end{itemize}
\State Reset \texttt{grid} state.
\While{\texttt{grid} episode not complete}
    \State Reset \texttt{market} state.
    \While{\texttt{market} episode not complete}
        \State \texttt{agents} submit bids, offers.
        \State \texttt{market} dispatches prices and quantities.
    \EndWhile~(\texttt{market} clears)
    \State \texttt{agents} submit \texttt{grid} actions.
    \State \texttt{grid} updates state.
\EndWhile
\end{algorithmic}
\end{algorithm}

From an object-oriented programming perspective, it is redundant to explicitly create an object to represent this environment if it simply iterates over all other objects. However, the environment object can provide the user with a much cleaner and easier-to-use interface. To highlight this, we show in the snippet of code below how to set up and simulate an environment having previously defined \texttt{grid}, \texttt{market} and [list of] \texttt{agents} according to the previous sections.

\vspace{0.75em}
\begin{tcolorbox}
\begin{minted}[fontsize=\footnotesize]{python}
env = Environment(grid, market, agents)
env.reset()
for t in env.iterate():
    pass
\end{minted}
\end{tcolorbox}
\vspace{0.75em}

We go into further details in the code's documentation about the different capabilities afforded by this style of interaction with the environment, such as the use of callbacks for example to easily save or extract data mid-simulation.

As shown in Figure \ref{fig:communication}, the environment can be accessed by all other objects and can access each of them. The figure appears to suggest that observation and control are proposed to be centralized. Indeed, we propose that in simulation, all objects should be able to access one another via a central \textit{Environment}, whereas in practical implementation there should be explicit communication networks that restrict this.

Our justification for this is that during simulation, ultimately there is only one user, which is the person using this platform. The best experience for them involves being able to easily access everything, as enabled by this `centralized' environment object. However, with that we must emphasize that we do not enforce that the user's implementation of the \textit{Grid}, \textit{Market} and \textit{Agent} objects satisfy the user's desired outcomes. Simply put, by providing flexibility under the framework, we hand over the responsibility to the user to ensure they are implementing the simulation the way they want to. Nonetheless, as shown in the section which follows, we provide templates that can aid users in designing their own use cases.

\vspace{1em}
\section{Use Cases}
\label{sec:Use_Cases}
In the previous section, we provided a blueprint for setting up and executing distribution electricity market simulations. In this section, we provide a few use cases of OpenGridGym for two main purposes: 1) to express the variety of simulations that could be executed, both in complexity and in relation to real-world problems, and 2) to show how OpenGridGym can be used to test and train AI-based approaches to solving problems in this domain.

Furthermore, we wish to convey to the reader that it is the simulation platform and its capabilities which we seek to highlight, rather than the use cases themselves. Neither the market mechanisms used nor the agent behaviors assumed in the use cases are suggested to be ideal. We provide those simplified use cases to reflect how users might interact with OpenGridGym to investigate the impact of possible designs and mechanisms that they may desire to experiment with.

\subsection{Use Case 1: Topology-Induced Market Power}

Here are the three goals of this use case:
\begin{itemize}
    \item Investigate the impact of network constraints in distribution grids on electricity pricing in a competitive market.
    \item Explore price-responsive elastic demand as an antidote to network congestion.
    \item Showcase the use of OpenGridGym's framework in setting up and executing a simulation environment to enable such inquiries.
\end{itemize}
To set up this use case, we follow the blueprint described in the previous section. That is, we first define the underlying model of the physical grid. Second, we suggest a market mechanism for dispatching prices and quantities based on consumer/producer input. Third, we introduce different kinds of agents to participate in this market, and describe their behavior. Finally, we show the results of the simulation extracted from the environment object.

\subsubsection{Grid Model}
We rely on the standard IEEE 34 node feeder as one specific model of a physical grid, and use OpenDSS to simulate power flow corresponding to different load and generation profiles. This distribution grid model is represented by a 69 kV feeder, a substation transformer which step-down to 24.9 kV, a mixture of single- and multi-phase unbalanced loads (28 in total), single- and three-phase distribution lines, capacitors and regulators.

OpenDSS offers the capability to connect directly to it using Python, and we supplement that with a user-friendly interface for the user to more seamlessly interact with it. That is, when agents submit what we referred to as `grid actions' in this paper, OpenDSS is called to solve power flow and return the results corresponding to those actions.

For the purposes of this specific use case, we are more interested in how agents interact in the market, rather than with the physical grid. Therefore, we reduce the grid model here to single-snapshot power flow, as opposed to a time-series.

\subsubsection{Market Mechanism}
In today's distribution grids, most consumers of electricity participate in local retail markets where a regional load serving entity offers them prices for consumption that could remain fixed for months or even years. With the increase of power demand on distribution grids, especially now with the introduction of fast-charging electric vehicles, there needs to be either an increase in the grid's capacity or some form of demand response to counteract that.

We explore a market mechanism which models consumers as price-responsive. In this model, consumer agents submit elastic demand curves as bids to the market, and the market is expected to dispatch demand corresponding to the submitted bids to ensure, using a simplified DC power flow model, that none of the distribution lines overflow. Let $q_{\text{bus}_i}$ represent net quantity consumed at bus $i$, and let $q_{\text{line}_{i,j}}$ represent the quantity of power flowing from bus $i$ to bus $j$. A consequence of radial topology is the following equality.
\vspace{0.5em}
\begin{align}
    q_{\text{line}_{i,j}} = q_{\text{bus}_j} + \sum_{k\in \mathcal{C}(j)} q_{\text{line}_{j,k}}
\end{align}
\vspace{0.2em}
where $\mathcal{C}(j)$ is the set of buses `children' to $j$ (i.e. downstream from the feeder). Knowing the topology, the market optimizes over all quantities consumed/produced in attempt to enforce thermal line limit constraints.

The market object is expected not only to dispatch the amounts of power consumed or produced by each participant, but to also dispatch electricity prices. Due to page limits for this paper, we defer the detailed formulation of the price-quantity optimization problem to a separate document, but here is the short summary. Consumers submit bids as demand curves, producers submit offers as supply curves, and for each of those, we rely on a simplified parametric model to define the curves. Supply and demand curves are chosen as affine relationships between price and quantity as follows ($s$ for supply, $d$ for demand):
\vspace{0.5em}
\begin{align}
    p_s &= \cfrac{p_\text{max} - p_\text{min}}{q_\text{max} - q_\text{min}}\left(q_s - q_\text{min}\right) + p_\text{min} \\
    p_d &= \cfrac{p_\text{min} - p_\text{max}}{q_\text{max} - q_\text{min}}\left(q_d - q_\text{min}\right) + p_\text{max} \\
    p_\text{max} &\geq p_\text{min} \\
    q_\text{max} &> q_\text{min}
\end{align}
\vspace{0.2em}
where $(p_\text{max}, p_\text{min}, q_\text{max}, q_\text{min})$ sufficiently parametrize either supply or demand curves. That is, for each bid or offer, an agent submits a quadruple of scalars to represent their participation in the market. Note that supply curves are strictly non-decreasing and demand curves are strictly non-increasing. Finally, it is the market's role to ensure that all participants are minimally satisfied while leaving no money on the table, expressed as follows:
\vspace{0.5em}
\begin{align}
    \left(p_s, q_s\right) &\text{ on or above supply curve} \\
    \left(p_d, q_d\right) &\text{ on or below demand curve} \\
    \sum_{d} p_d q_d &= \sum_{s} p_s q_s
\end{align}
\vspace{0.2em}


The objective used in the market's optimization problem is to maximize the sum of net consumer and net producer surpluses. We know from classical economics that in cases without any network constraint, i.e. no limit on exchange of quantity between supply and demand, the optimal price-quantity dispatch is the intersection of aggregated supply and demand curves. However, in case of network constraints, the optimal solution becomes less trivial. The purpose of this use case is not to advocate for this market mechanism, but rather to test the impact of demand and supply flexibility (or elasticity) on consumer prices. We say a consumer is less \textit{price-responsive} the steeper their demand curve is.

\subsubsection{Agent behavior}
To compare price-responsive to price-unresponsive consumers, we use the parametrization of demand curves described in the market formulation. Each consumer agent submits a quadruple of scalars $(p_\text{max}, p_\text{min}, q_\text{max}, q_\text{min})$ provided that $p_\text{max} \geq p_\text{min}$ and $q_\text{max} > q_\text{min}$, and in doing so, they declare their bid to the market. Producer agents do the same with parametrized supply curves to declare their offers to the market.

Each agent seeks to maximize their own personal rewards, which are made up of two components. For each price-quantity dispatch from the market, supply agents receive a net amount of money equal to the product of price, payed for by the demand agents. Furthermore, for each supply agent, there is an opportunity cost $c_s(q_s)$ for each quantity $q_s$ dispatched by the market. This quantity is hidden from everyone but the agent. Similarly, for each demand agent, there is a hidden utility $u_d(q_d)$ in consuming dispatched quantity $q_d$. From these quantities, we define the market dispatch-based reward signal $r$ for each agent as follows:
\vspace{0.5em}
\begin{align}
    r_s\left(p_s, q_s\right) &= p_s q_s - c_s(q_s) \\
    r_d\left(p_d, q_d\right) &= u_d(q_d) - p_d q_d
\end{align}
\vspace{0.2em}

Economics theory suggests that in purely competitive markets, it is in the interest of producers to submit supply curves that most closely reflect their `true' opportunity costs, to remain competitive. When faced with network constraints however, it is possible for producers with market power to capitalize and submit much higher prices than their cost, which we explore in this use case. Nonetheless, in general, it is left to the user to utilize Python's AI-friendly ecosystem to train this agent to maximize its rewards.

\subsubsection{Environment Results}
For this use case, we initialize 28 demand bids (one per load) as inelastic, and we introduce a total of 5 supply units with equal supply capacity, spread uniformly across the grid. One supply unit is located at the feeder node of the distribution network with a horizontal supply curve, i.e. fixed price for all quantities, derived from transmission-level wholesale, and assumed to be exogenous to our simulation environment. The other four supply units are price-responsive, i.e. their supply curves are increasing. Under this scenario, we observe that all consumers are dispatched a price of 4.3 ¢/kWh with no congestion in the network.

\begin{figure}[!ht]
\centering
\includegraphics[width=0.45\textwidth]{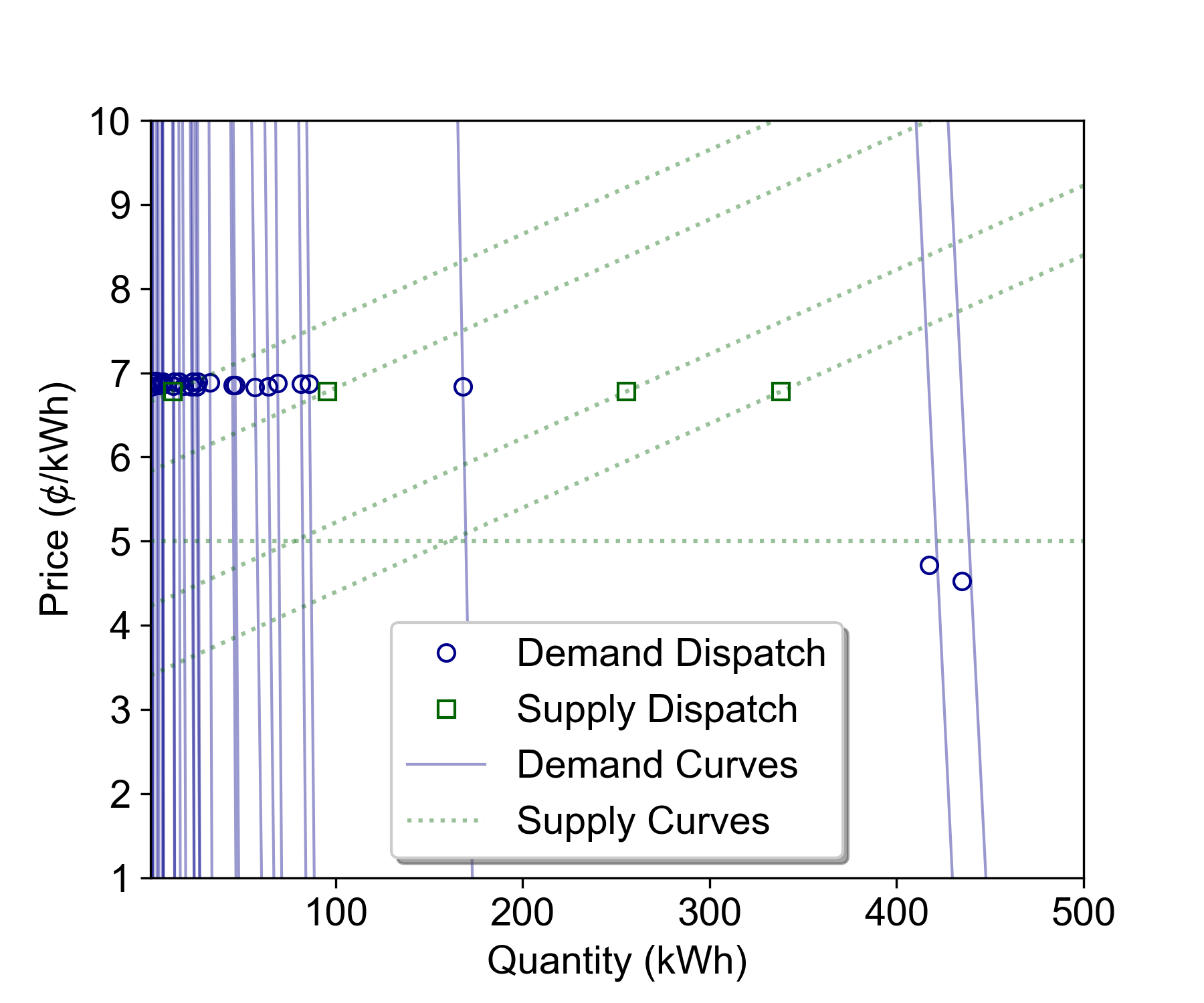}
\caption{Market dispatch for inelastic demand and elastic supply. Prices increase for most consumers due to line congestion.}
\label{fig:inelastic_gen_pricey}
\end{figure}

Next, we iterate through the market, where agents update their bids and offers, and we purposefully instruct consumer agents to keep their demand inelastic. The results are shown in Figure \ref{fig:inelastic_gen_pricey}. The natural consequence is that prices must go up since generators increased their price offers with no objections. The figure shows us two things. First, we see that not all consumers now pay the same price (\{4.6, 4.8 and around 6.9\} ¢/kWh). Second, even though the feeder node remained fixed price (and cheaper than other supply), consumers still buy at more expensive prices, simply due to network congestion.

Since the network is congested, it limits the amount of cheap power deliverable to certain pockets of load in the network, thus creating an advantage to local pricey generators. This is a fundamental property of congested radial networks. We do not wish to go deeper in this use case to analyze this market mechanism. However, we'd like to continue this example to show how consumer price-responsiveness can serve as an antidote to this topology-induced price increase.

\begin{figure}[!ht]
\vspace{-1em}
\centering
\includegraphics[width=0.45\textwidth]{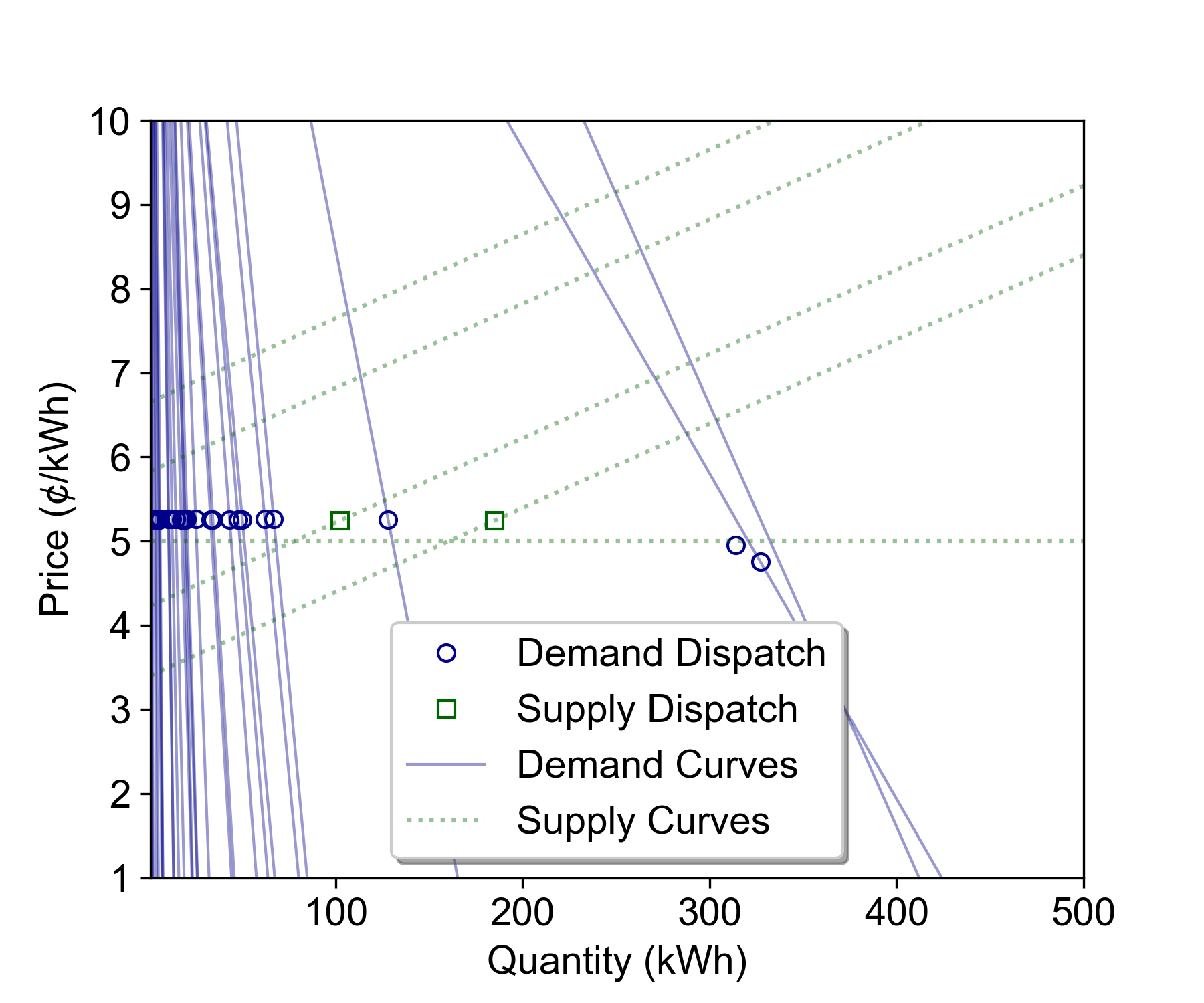}
\caption{More elastic demand curves lead to lower prices.}
\label{fig:inelastic_gen_pricey_load_responsive}
\vspace{-1em}
\end{figure}

Proceeding in the simulation, supply agents still submit relatively high prices, but the demand agents now submit more elastic demand curves. As a result, the market dispatches lower consumption quantities, even though the network remains congested. This is shown in the updated supply/demand curves of Figure \ref{fig:inelastic_gen_pricey_load_responsive}.

\begin{figure}[!ht]
\centering
\includegraphics[width=0.49\textwidth]{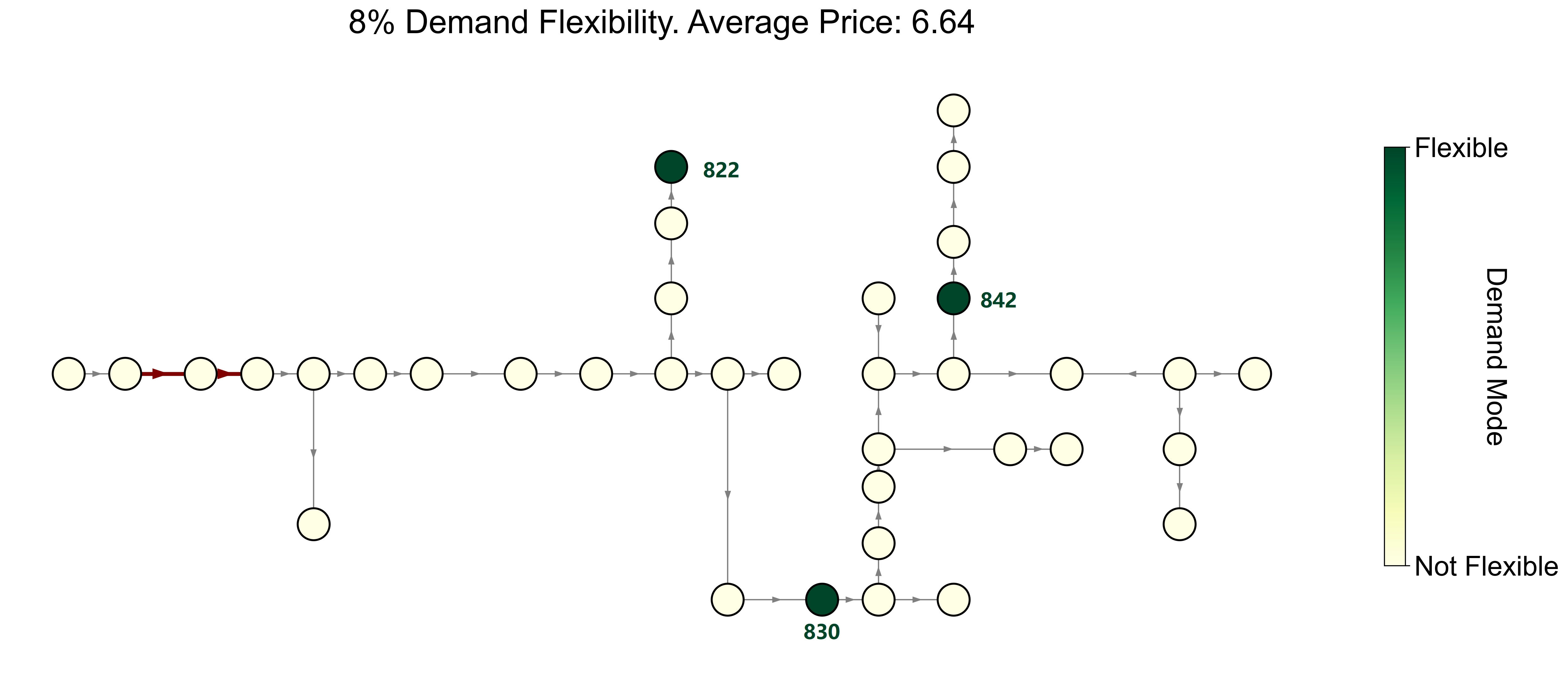}
\includegraphics[width=0.49\textwidth]{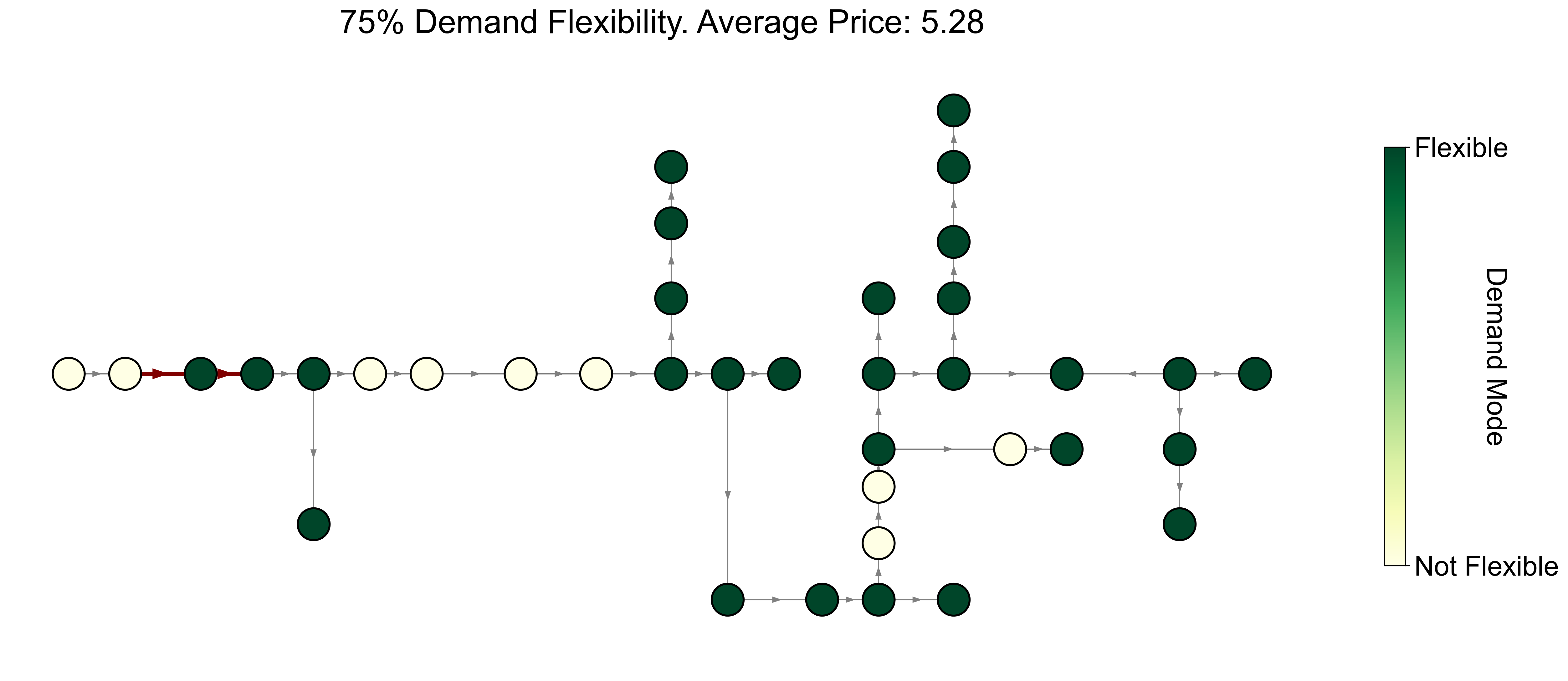}
\caption{Average consumer price for a congested network with different levels of demand flexibility.}
\label{fig:grid_load_responsive}
\end{figure}

A natural question that may arise is, how much demand flexibility is needed to lower the prices? To answer this, we do not change supply agents' offer for now. We just randomly select bid offers by demand agents such that some of them are flexible, and some not. We consider two scenarios, one with 8\% flexible, and one with 75\%. As shown in Figure \ref{fig:grid_load_responsive}, for the former case, the average consumer price is 6.64 ¢/kWh, whereas for the latter, the average price is 5.28 ¢/kWh.

To conclude this use case, here are three takeaway messages we wish to deliver. First, by following OpenGridGym's framework to set up this use case, we can very easily separate and debug each of the grid, market and agent simulation objects. Second, by loading a pre-defined OpenDSS-based IEEE test case, we didn't need to create the actual network, rather just specify it. This means the user can repeat this experiment on different standard grids easily, or of course create their own. Third, we were able to simulate a meaningful experiment using this framework. Namely, in this case, we were able to show that even in congested radial networks, via demand flexibility you can reduce the prices by some margin. Granted, this specific market mechanism calls for much more in-depth exploration, but here, the purpose is to provide a template for users to get an idea of how OpenGridGym works.

\subsection{Use Case 2: Learning in Peer-to-Peer Markets}
Consider a market of the sharing economy \cite{xia2019small, sharing_economy} where market participants can switch roles between producers and consumers depending on availability of their resources. For example, PV home owners generating extra energy could shortly serve as producers by providing additional resources to the grid and they could also serve as consumers in this market when they do not produce energy from PVs. Thus, market participants are turned into prosumers who can both produce and consume resources. In this use case, we model such PV home owners as agents and consider the Peer-to-Peer (P2P) market. Here, geographic locations and weather determine which agents shift from being producers to consumers; If there are no available PV home producers, consumers can also buy the electricity from a utility provider. In this market, the role of the market maker is to match the producers and consumers so that a producer-consumer pair can negotiate to make a transaction if agreed. The objective of the market maker is to serve as a middleman between producers and consumers while respecting physical grid constraints. In this market, agents are concerned with a simple collection of tasks that they negotiate (bargain) the \textit{agreed} price with other agents.

\subsubsection{Grid Model}
We use the same grid model used in Use Case 1.

\subsubsection{Market Mechanism}
The role of this P2P market is to match producers and consumers so that a producer-consumer pair goes through a number of negotiation steps to find an agreement on the price. If they both \textit{agree} on the price, trade happens; otherwise they fail to make a trade. To be specific, the market maker allows a matched pair of producer and consumer to bid a number of negotiation steps, denoted by $T \in \mathbb{N}$. For each market negotiation step $t < T$, an agent (either producer or consumer) can explore to learn an optimal action, which depends on the opponent's strategies and vice versa. Given a matched pair, denote bid offers from producer and consumer by $b_{p,t}$ and $b_{c,t}$ respectively at time $t$. If $b_{p,t} \le b_{c,t}$, the market maker says a trade is successful. Producer $p$ receives a reward $r_{p,t} = b_{p,t} - c_{service}$ and consumer $c$ receives a reward $r_{c,t} = ub - b_{p,t} - c_{service}$ where $c_{service}$ is a fixed service fee in the market and $ub$ is an utility price that consumer can alternatively purchase from. If $b_{p,t} > b_{c,t}$, then both producer and consumer receive a reward $r_{p,t} = r_{c,t} = -c_{lose}$ as a penalty when the transaction is not successful.

Lastly, we note that the market performs random matching between producers and consumers without considering agents' locations, generation amount or consumption demand.

\subsubsection{Agent behavior}

We make a number of assumptions about agent behavior for simplicity of demonstration. First, we assume that all producers are homogeneous and generate the same fixed amount of electricity, say 3 kWh, and consumers purchase the same quantity. If this assumption is not met, the market maker can do a maximum weight matching or other alternative matching algorithms with respect to generation amount and demand. When agents are matched by the market maker with an opponent and negotiate, we assume that they have finitely many arms (options) to choose from in their bid offers. Finally, we assume that all agents implement the Upper Confidence Bound (UCB) algorithm \cite{lattimore2020bandit}. That is, when agents are matched by the market, a matched pair of producer and consumer take actions according to the UCB algorithm which is described as follows: Let $T_{i}(t-1)$ denote the number of samples from arm $i$ up to time $t-1$ and $\hat{\mu}_{i}(t-1)$ denote the empirical mean (received rewards) from that arm $i$ up to time $t-1$. Now, define $\text{UCB}_{i}(t-1,\delta)$ for each arm $i$ as follows:
\vspace{0.5em}
\begin{align}
    \text{UCB}_{i}(t-1,\delta) =
    \begin{cases}
        \infty & \text{if } T_{i}(t-1) = 0 \\
        \hat{\mu}_{i}(t-1) + \sqrt{\frac{2 \log(1/\delta)}{T_{i}(t-1)}} & \text{otherwise}.
    \end{cases}
\end{align}
\vspace{0.2em}
where $\delta$ is the error probability. At negotiation time $t$, each agent takes an action $a \in \arg \max_{i} \text{UCB}_{i}(t-1,\delta)$, updates the empirical mean $\hat{\mu}_{i}(t)$ for arm $i$, and repeats this negotiation steps until $t = T$. For details about the UCB algorithm and analytic results, we refer readers to \cite{lattimore2020bandit}.

\subsubsection{Environment Results}

Using the \textit{Grid}, \textit{Market} and \textit{Agent} formulation above, we implement the environment and demonstrate negotiations between a matched pair of producer and consumer. We show a multi-period negotiation between matched pairs. For each market step, the producer and consumer agents negotiate for a trade over the next several grid steps. At the time of the negotiation, generation and load forecast time series are used for the producers and consumers respectively. If a consumer cannot get enough energy from its paired producer for either supply deficiency or unsuccessful negotiation, the energy deficiency is cleared using grid power drawing from the substation at the retail price from the utility company. This formulation is able to account for the thermal inertia of loads across different market intervals and uncertainty of future renewable generation after the power transaction settlement. The reward and action trajectories for the multi-period case are shown below in Figures \ref{fig:usecase2_reward_traj_mp} and \ref{fig:usecase2_action_traj_mp}, both plotting the moving average of the past 200 iteration steps of agent actions and rewards. The reward trajectories for both producers and consumers in Figure \ref{fig:usecase2_reward_traj_mp} show that they are successful at negotiation and receive positive rewards, instead of negative rewards when negotiation is unsuccessful. This is evident from agents' action trajectories in Figure \ref{fig:usecase2_action_traj_mp} where consumers' actions are always higher than or equal to the producers' actions after around 2000 negotiations steps. Thus, the power transactions are successful and rewards are positive for agents in the market.

\vspace{-1em}
\begin{figure}[!ht]
\centering
\includegraphics[width=0.49\textwidth]{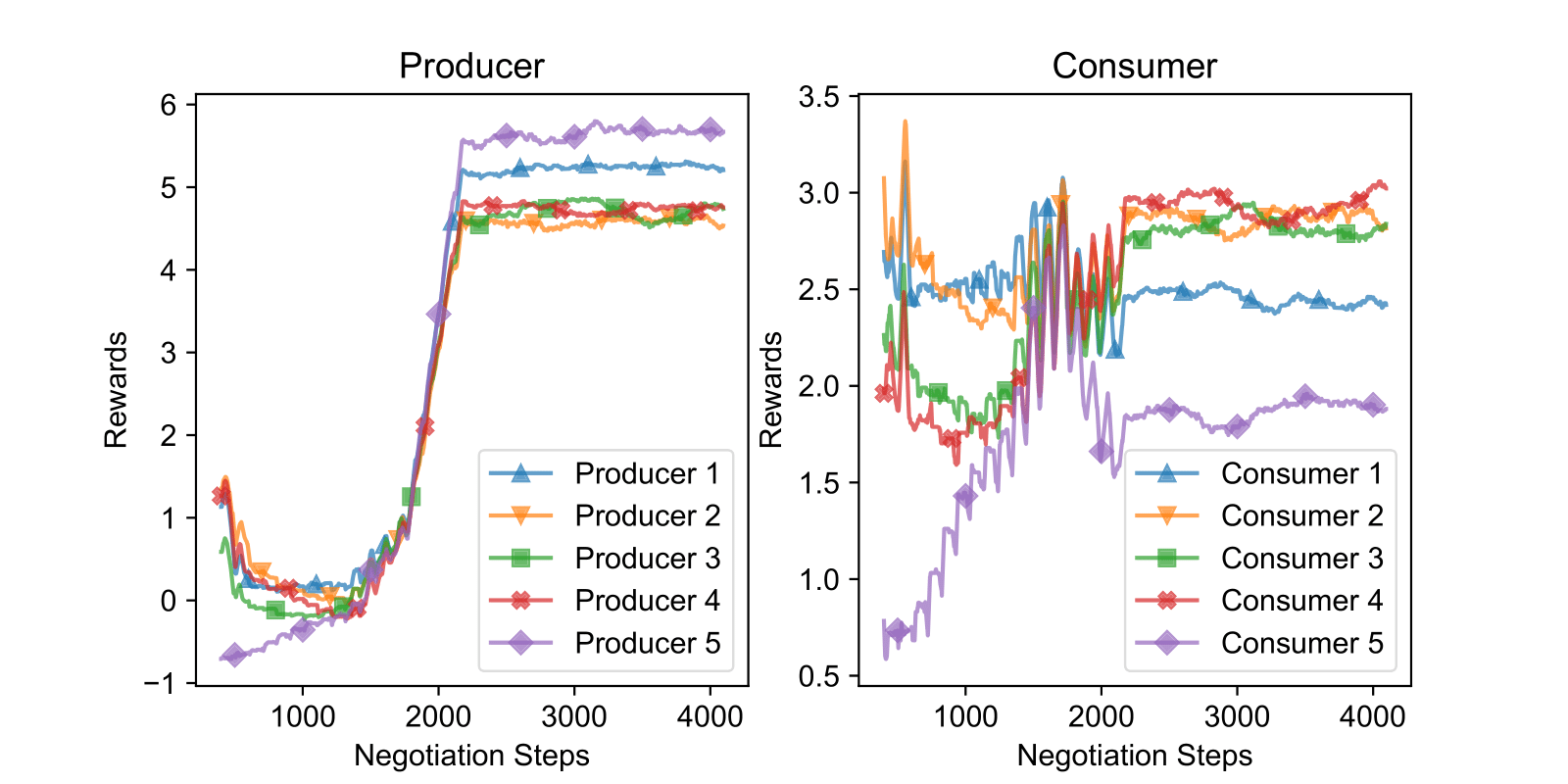}
\caption{Reward trajectories of producers and consumers.}
\label{fig:usecase2_reward_traj_mp}
\end{figure}
\vspace{-2em}
\begin{figure}[!ht]
\centering
\includegraphics[width=0.49\textwidth]{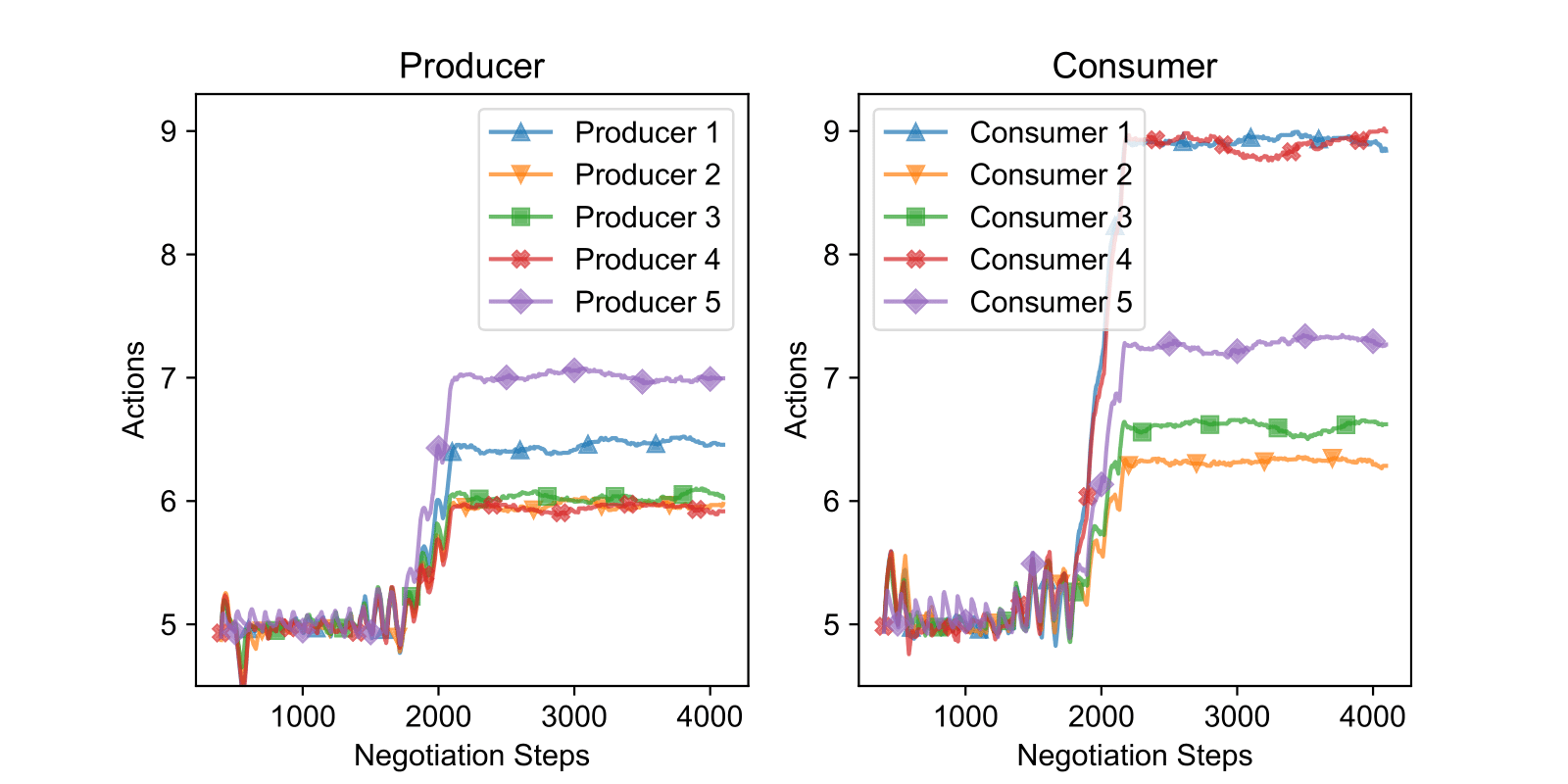}
\caption{Action trajectories of producers and consumers.}
\label{fig:usecase2_action_traj_mp}
\end{figure}
It is also noticed in numerical experiments that under certain operating conditions the settled bilateral transactions cannot be realized in the grid step. In radial distribution networks there is usually one electric path between producers and consumers at different buses, when the power being traded exceed rating of any line in the grid model the power flow results become infeasible. Under such conditions, market transactions can potentially impact the operation of physical distribution grid.

\subsection{Use case 3}
The third use case attempts to establish a general comprehensive market structure for future distribution systems where prosumers in the network trade their power generation and consumption with fellow prosumers or with the transmission grid. There are three roles for any prosumer to participate in the market: as a power producer, power consumer or demand response provider. Power producers are able to inject net positive real power into the network with distributed energy sources such as PVs, diesel engines or charged batteries. Power consumers have net negative real power capacity that need to be supplied by the transmission grid or other producers. In the case of network congestion which results in high price or insufficient supply, demand response providers at certain parts of the network can voluntarily reduce their power consumption in exchange for a profit, which is paid for by other consumers that benefit from such reduction.

\subsubsection{Grid Model}
The IEEE 37-node feeder benchmark system is used in this case study. This system is a 4.8kV mid voltage distribution with around 2200kW total load. This network is characterized by a all-delta configuration with one large three-phase load and many spot single-phase loads. The system information is automatically extracted and processed using OpenGridGym.

\subsubsection{Market Mechanism}
Using our proposed framework, we demonstrate a possible market mechanism for distribution system operators (DSOs) that utilizes the concept of Distribution Locational Marginal Prices (DLMPs). As with LMP's in current transmission systems, power producers submit a convex cost function that maps their net power production to desired revenue. The DSO then runs security-constrained optimal power flow (SCOPF) to determine the exact dispatch for every producer in the network that minimizes the total energy cost. A distinctive feature in distribution systems compared to transmission systems is the existence of an infinite source. Most if not all distribution networks are connected to the transmission system through a transformer in a substation whose capacity rating is large enough to supply the entire distribution network. However, the price to use power from the transmission system can also change with time, as the price may be affected by the transmission system LMP. The detailed problem formulation of SCOPF is as follows:
\vspace{0.5em}
\begin{multline}
    \min_{P^G, P^D}   LMP^{source} max(0, P^{source}) + \\ c_{gen}(P^g) + c_{DR}(\overline{P^d} - P^d)
\end{multline}

$\text{subject to:}$\\
\begin{align}
    P^{source} = 1^T(P^d - P^g)\qquad \text{(power balance)} \\
    P^{g}_{min} \leq P^g \leq P^g_{max} \qquad \text{(generator limit)} \\
    0 \leq P^d \leq \overline{P^d} \qquad \text{(demand response)} \\
    \Delta p = \Delta P^g - \Delta P^d \qquad \text{(bus injections)} \\
    \Delta f^{line} = H \Delta p \qquad \text{(power transfer)} \\
    -f_{max} \leq f^{line} \leq f_{max} \qquad \text{(line flow limits)}
\end{align}
\vspace{0.2em}

where $c_{gen}$ and $c_{DR}$ are the cost functions of distributed generators and demand response providers; $P^{source}$ is the total power drawn from the substation transformer; $P^d$ and $P^g$ are the power consumption and generation at each bus, while $\overline{P^d}$ is the initial load value before demand response; $f^{line}$ is the real power flow in lines and must not exceed the ratings; $H$ is a matrix computed from the network topology that maps net real power injections at each bus to real power flows at each line.
    
\subsubsection{Agent behavior}
In a learning problems, the cost curves (which may be parameterized as polynomial of piecewise functions) of power producers and demand response providers are determined by agents. During every market clearance interval, each agent observes market information disclosed by the DSO and determine the parameters of its cost curve, $c_{gen}$ or $c_{DR}$. After the DSO solves the SCOPF problem, each agent receives a dispatch that specifies how much power they are allowed to produce and consume. The reward of each agent is then calculated based on their net energy consumption, production or reduction and the DLMP at their bus.

\subsubsection{Environment Results}
\begin{figure}[!ht]
\centering
\includegraphics[width=0.49\textwidth]{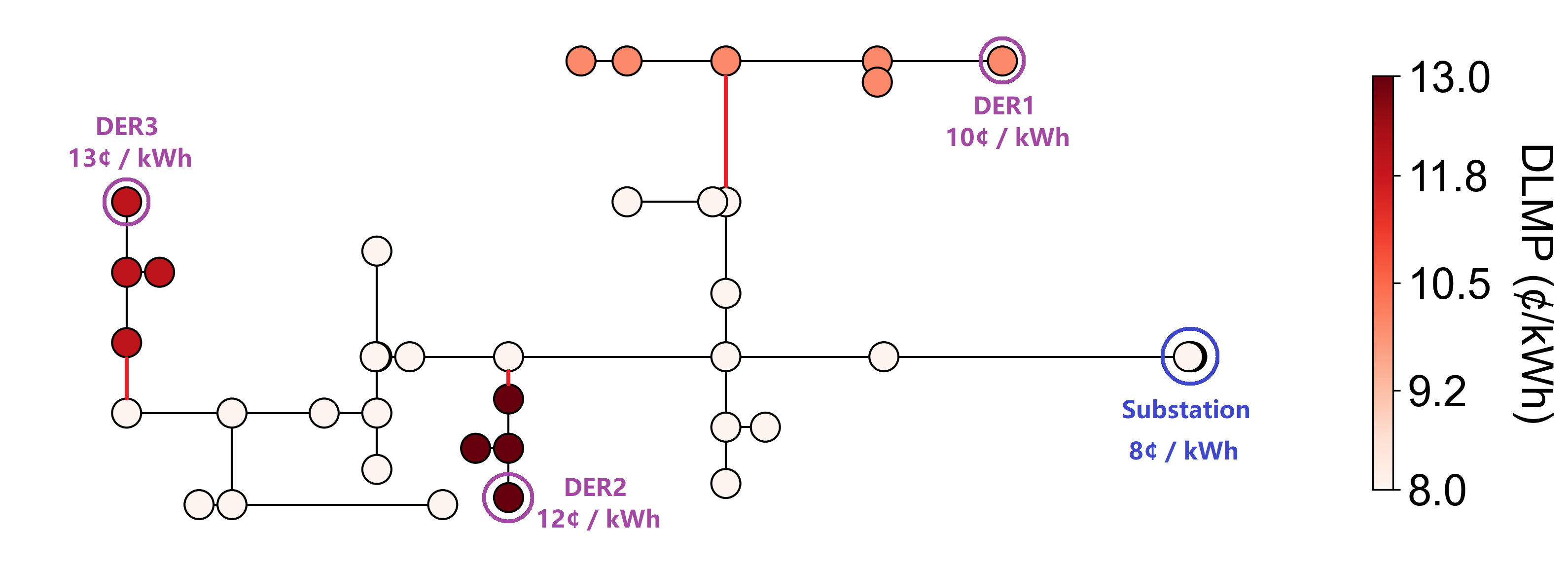}
\caption{Distribution of DLMPs in the IEEE 37-Node Feeder}
\label{fig:usecase3_lmp}
\end{figure}

OpenGridGym provides an implementation using CVXPY \cite{CVXPY} and NetworkX \cite{networkx}. At each market step, the DSO uses the agents' input to compute a DLMP for every load bus in the network, which is then used to calculate the cost and revenue of all market participants. An example of DLMP distribution in a simple radial network is shown in Fig. \ref{fig:usecase3_lmp}.

\section{Concluding Remarks}
\vspace{0.5em}
\label{sec:Conclusion}
This paper presents a simulation platform, OpenGridGym, for scalable multi-agent market simulation for future distribution systems. This platform is open-source and based on user-friendly Python toolkit. This platform could serve as a benchmark for the research community to simulate and analyze the outcome of various market mechanisms with direct access to AI-friendly ecosystem via Python. 

Building upon this open-access toolkit, many interesting research questions in distribution market design and outcome could be quantitatively analyzed. To showcase OpenGridGym, we present different use cases and demonstrate how users can easily integrate trainable AI-driven agents into their simulation. Future work on OpenGridGym includes expanding the use cases to provide users with even more templates to follow, and to assist the design of alternative market mechanisms to address challenges faced in modern distribution grids.


\ifCLASSOPTIONcaptionsoff
  \newpage
\fi



\vspace{1em}
\bibliographystyle{IEEEtran}
\bibliography{references.bib}

\end{document}